\documentclass[5p,times,twocolumn,authoryear]{elsarticle}

\usepackage{amssymb}
\usepackage[authoryear]{natbib}

\usepackage{hyperref, booktabs}
\usepackage{amsmath}
\usepackage{threeparttable}
\usepackage{cleveref, multirow, multicol}
\usepackage{xspace}
\usepackage{xcolor}
\newcommand{\onedot}{.\xspace}
\def\@onedot{\ifx\@let@token.\else.\null\fi\xspace}
\def\eg{\emph{e.g}\onedot} 
\def\ie{\emph{i.e}\onedot} 
 
\def\etc{\emph{etc}\onedot} \def\vs{\emph{vs}\onedot}

\journal{Nuclear Physics B}

\begin{document}

\begin{frontmatter}

\title{ObliCity: A Benchmark and Baseline for Roof-to-Ground Projection Displacement Correction} %% Article title

\author[UCAS,CityU]{Kai Li}
\ead{likai211@mails.ucas.ac.cn}

\author[AIR]{Yupeng Deng\corref{cor1}}
\ead{dengyp@aircas.ac.cn}
\author[UCAS]{Ligao Deng}
\author[AIR]{Zhihao Xi}
\author[UCAS]{Chenhao Wang}
\author[HKU]{Jierui Zhang}
\author[UCAS]{Yingrui Ji}
\author[AIR]{Yu Meng}
\author[CityU]{Xiangyu Zhao}

\cortext[cor1]{Corresponding author}
%% Author affiliation
\affiliation[UCAS]{organization={School of Electronic, Electrical and Communication Engineering, University of Chinese Academy of Sciences},%Department and Organization
            addressline={1 East Yanqi Lake Road}, 
            city={Beijing},
            postcode={100049}, 
            state={China}}
\affiliation[CityU]{organization={College of Computing, Department of Data Science, City University of Hong Kong},%Department and Organization
            addressline={Kowloon Tong}, 
            city={Hong Kong},
            postcode={999077}, 
            state={China}}
\affiliation[AIR]{organization={Aerospace Information Research Institute, Chinese Academy of Sciences},%Department and Organization
            addressline={9 Dengzhuang South Road}, 
            city={Beijing},
            postcode={101408}, 
            state={China}}
\affiliation[HKU]{organization={Department of Electrical and Computer Engineering, The University of Hong Kong},%Department and Organization
            addressline={Pokfulam}, 
            city={Hong Kong},
            postcode={999077}, 
            state={China}}

\begin{abstract}
Oblique-view urban remote sensing imagery inevitably exhibits geometric projection displacements between building roofs and footprints, leading to significant distortions in spatial structure. Existing approaches either ignore these deformations or handle them implicitly within segmentation-based frameworks, where progress is dominated by general segmentation advances rather than improvements in geometric correction.
In this work, we explicitly define roof-to-footprint offset vector (RFOV) extraction as an independent learning task that decouples geometric alignment from semantic segmentation. To support this task, we introduce the Oblique City dataset (ObliCity), the first large-scale benchmark that integrates high-resolution UAV imagery and globally distributed satellite data, covering diverse city morphologies and camera perspectives.
Methodologically, we reformulate DragOSM into DragRoof, an ODE-based framework inspired by human annotation behavior. By simulating the continuous process of dragging roofs toward their footprints, DragRoof learns deterministic, geometry-consistent offset fields and adaptively determines convergence through an end token.
Extensive experiments on ObliCity demonstrate that DragRoof achieves state-of-the-art RFOV extraction performance, requiring fewer inference steps while delivering superior directional and length accuracy. Our dataset and model establish a principled foundation for studying projection displacement correction in oblique remote sensing imagery. The source code and dataset will be avaliable at \url{https://github.com/likaiucas/DragRoof}.
\end{abstract}

%% Keywords
\begin{keyword}
Offset learning, building extraction, off-nadir remote sensing image.

\end{keyword}

\end{frontmatter}

\section{Introduction}
\label{sec:intro}
\begin{figure}
    \centering
    \includegraphics[width=\linewidth]{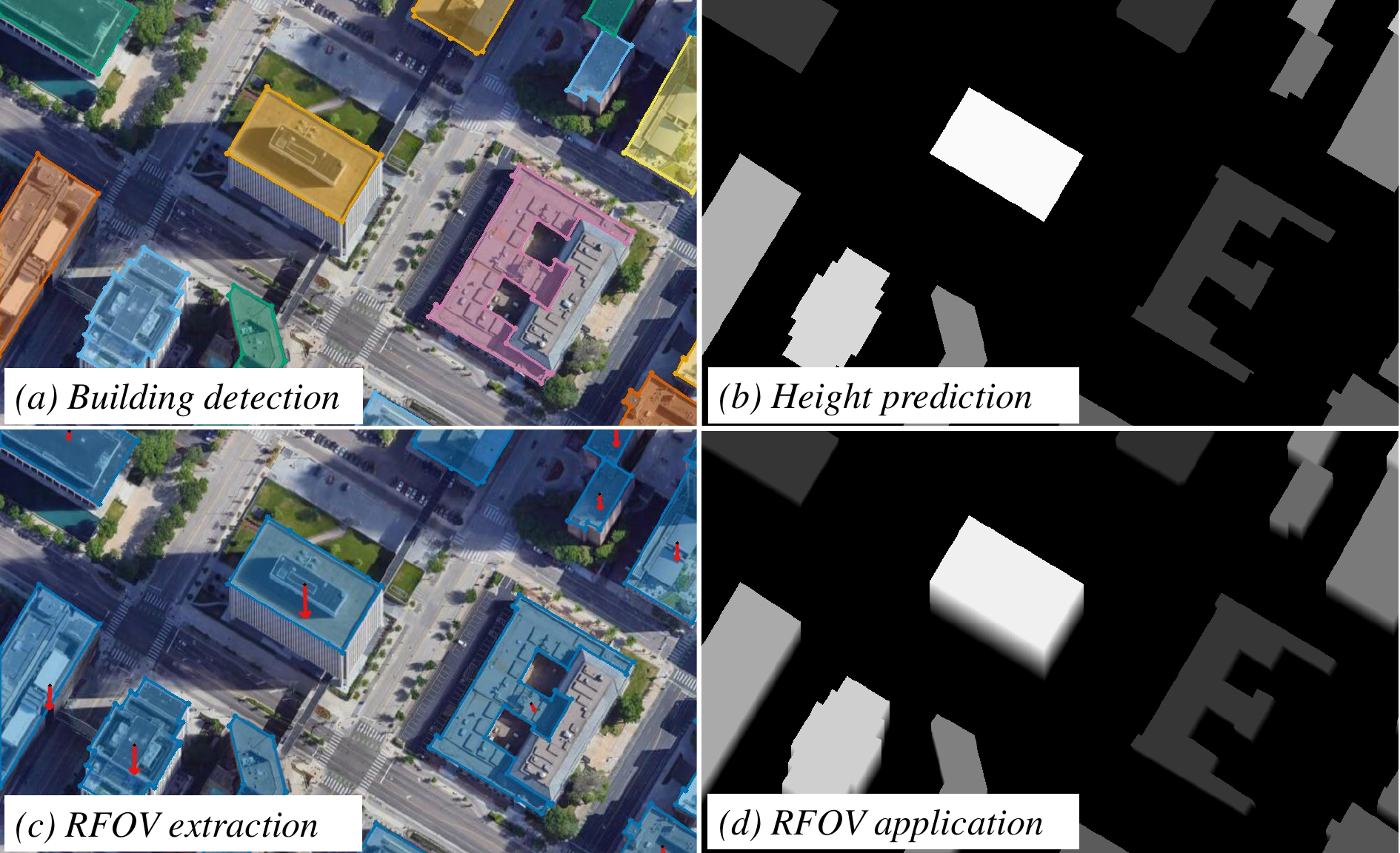}
    % 当下针对建筑物的主流研究是building detection和height prediction. (a) Building detection的研究方向则以建筑物的语义特征的提取为主，\eg, roof的instance segmentation与classification. 而建筑物的height prediction (b)则又过分依赖于深度学习中的方法，让模型直接为建筑物的局部语义预测一个明确的物理高度（米）. （c）我们提出的任务是针对给定影像中的建筑物提取RFOV（黄色vector）。（d）基于建筑物的RFOV特征，我们可以更好地描绘建筑物的结构、位置等，而又结合遥感影像的像素分辨率以及相机状态可知的特点，使用RFOV转化得到的建筑物像素高度相比于（b）而言会更加physically reliable。
    \caption{Current mainstream research on buildings mainly focuses on two directions: (a) {Building detection}, which emphasises semantic feature extraction, \eg, roof instance segmentation and classification; and (b) {Height prediction}, which often relies heavily on deep learning models to directly regress physical height values (in meters) from local semantic cues.  
(c) We propose a new task that extracts the RFOV (red arrows) for each building in a given image.  
(d) Based on the RFOV features, we can more accurately describe building structures and positions; and, by leveraging the known pixel resolution of remote sensing imagery and camera geometry, the pixel-level height derived from RFOVs is more {physically reliable} than direct height regression in (b).
}
    \label{fig:intro}
      \vspace{-5mm}
\end{figure}

Urban remote sensing images captured from oblique views inevitably suffer from geometric projection displacements between building roofs and their corresponding footprints~\citep{bfe_review}. These roof-to-footprint displacements, caused by camera tilt and building height, distort the spatial structures of buildings and make it difficult to obtain accurate building geometry from monocular imagery. 
Currently, building detection and height estimation are two of the most prominent research directions.
Existing mainstream building extraction studies, however, predominantly focus on roof segmentation~\citep{hisup, sampolybuild, vit_building_extraction, P2PFormerV2} (Fig.~\ref{fig:intro}(a)), while largely overlooking such projection-induced deformations. Many of these approaches treat the roof as a proxy (implicitly assuming roof–footprint perfectly overlapped, which no longer holds in modern ultra-high-resolution imagery) for the entire building or employ semantic segmentation to estimate per-pixel building height values~\citep{cao2021deepheight, height_tgrs, sun2024gableheight, Gultekin_2025_ICCV_height} (Fig.~\ref{fig:intro}(b)). Although effective for general building delineation, these representations (roof instances or semantic height maps) are insufficient to recover precise ground-level positions or the 3D structure of buildings. % 图片展示

% 尽管在一些针对offnadir image的building footprint mask extraction任务中，RFOV开始做为roof segmentation的辅助任务，以帮助获得最终的footprint. 此类方法的创新依然聚焦于semantic segmentation的发展, 因为footprint mask extraction任务的评估聚焦于使用一些常见的mask指标，\eg, F1score or \textit{Intersection over Union} (IoU). 这些指标对于RFOV是不敏感的，[因为]在本身roof mask的轮廓预测不够精确的情况下，RFOV即使预测的精度提高3-5个像素，在mask metrics的反馈也不显著。这种情况特别针对于roof mask较大，而RFOV较小的数据，尤为显著。因此，RFOV提取的技术进步往往因为roof mask提取技术的进步而被忽视。
Although RFOVs have been introduced in several off-nadir building footprint extraction tasks as auxiliary cues to assist roof segmentation~\citep{BONAI, pang2023detecting, MTBRNet, MLS-BRN, obm, polyfootnet,zhou2025nesf}, the innovations of these methods still primarily focus on advances in semantic segmentation. 
This is because the evaluation of footprint extraction relies on common mask-based metrics, \eg, F1 score and \textit{Intersection over Union}, which are largely insensitive to RFOV accuracy\footnote{We mathematically probe this in Sec.~\ref{supp:rfov_mask}.}. 
When the roof mask itself is not precisely delineated, even an improvement of 3–5 pixels in RFOV prediction yields negligible gains in mask metrics. 
This issue is particularly evident in cases where the roof mask is large but the RFOV magnitude is small. 
Consequently, progress in RFOV extraction has often been overshadowed by advances in roof segmentation performance. 

This motivates us to decouple RFOV extraction from conventional segmentation frameworks, framing it as a standalone learning objective. \ie, assuming that reliable building regions (roofs) are available as priors, we only estimate an RFOV for each building, as Fig.~\ref{fig:intro}(c). Within these task settings, the future technical improvement on the RFOV extraction will benefit downstream applications (Fig.~\ref{fig:intro}(d)). 

% However, 当下包含RFOV标签的数据集主要是BONAI和OmniCity, 然而当前的数据集面临着场景缺失（BONAI仅使用0.3-0.5m的中国5个大型城市的卫星影像）和偏移量不显著（OmniCity约60%的RFOV不足10像素）的问题. 同时，而作为普遍存在于UAV影像中的ultra-long RFOV难样本数据是稀缺的，且在模型层面难以精准预测其长度的。
However, existing datasets containing RFOV annotations are limited to BONAI~\citep{BONAI} and OmniCity~\citep{li2023omnicity}. 
Both suffer from significant shortcomings: BONAI includes only satellite imagery from five major Chinese cities with 0.3–0.5\,m resolution, while over 60\% of the RFOVs in OmniCity are shorter than 10 pixels, making displacement cues weak. 
Moreover, ultra-long RFOV samples, which are commonly present in UAV imagery, remain scarce, and existing models still struggle to accurately predict their lengths. On the other hand, existing models capable of directly predicting RFOVs are mainly represented by OBM~\citep{obm} and PolyFootNet~\citep{polyfootnet}. 
These methods were originally designed to facilitate the manual annotation of building footprints, and thus adopt simple yet spatially ambiguous bounding boxes as prompts for human–model interaction.

% To address the aforementioned gaps, 我们对珠三角地区的七个不同区域的UAV影像进行了精细化的标注，同时对开源矢量数据集IRSAMap的城市部分进行了重新标注，构建了首个涵盖多场景、多卫星平台的数据集ObliCity。同时，受启发于数据集的人类标注过程，通过获取 沿着建筑物facade的轮廓边缘将roof拖拽到footprint位置直到与局部可见的像素重合 的轨迹，而获取RFOV。
To address the aforementioned gaps, we meticulously annotated UAV imagery (0.1\,m) from seven distinct regions across the Yangtze River Delta and re-annotated the urban subset of the open-source vector dataset IRSAMap~\citep{meng2025irsamap}. 
Together, these efforts constitute the {ObliCity} dataset, the first multi-scene, multi-platform dataset that integrates UAV and satellite imagery for RFOV extraction.

On the methodological side, inspired by the human annotation process in {ObliCity}, we model RFOV extraction as a continuous process that mimics {how annotators drag the roof along the building facade boundaries toward its footprint until the projected edges align with visible image pixels}. 
We formulate this trajectory as an ordinary ODE-based\footnote{Ordinary Differential Equation (ODE)} inference process and instantiate it within the DragOSM framework; we name this model as DragRoof. 
Furthermore, to enable the model to intuitively determine whether the dragging process has reached the footprint position, we introduce an \textit{end token} that serves as an indicator of convergence. 
DragRoof uses iterative inference to handle ultra-long RFOV cases. 

% Specifically, DragRoof在训练的过程中首先从0到1的均匀分布采样时间$t$，用来模拟人类标注过程中roof在facade区域内的中间状态，然后训练DragRoof去拟合当前采样时间的位置到footprint位置的向量，并通过end token解码一个end flag，该end flag与档mask输入位置到footprint的距离高度绑定，用以感知预测的重点。在推理的过程中，DragRoof通过连续校正roof mask的过程，获得最终的RFOV。经实验，DragRoof仅需约2步的去噪过程，便实现了SOTA的RFOV提取。巧合的是，通过归纳DragOSM为SDE维纳过程建模，从DragOSM需要约5步的去噪过程到DragRoof的两步高效去噪恰好符合了图像生成领域从DDPM到Flow Matching的技术进步。
Specifically, during training, DragRoof uniformly samples a time step $t \in [0,1]$ to simulate the intermediate states of the roof being dragged along the facade, analogous to the human annotation process. 
The model is then trained to regress the displacement vector from the current sampled position to the footprint, while the end token decodes an \textit{end flag} that is strongly correlated with the distance between the current mask and the footprint, allowing the model to sense convergence. 
During inference, DragRoof progressively refines the roof mask through a continuous correction process to obtain the final RFOV. 
DragRoof achieves state-of-the-art performance with only two-step inference. 

Different from OBM and PolyFootNet, DragRoof performs RFOV extraction using roof polygons or mask prompts as inputs, enabling better compatibility and integration with other mask–based methods, \eg, HiSup~\citep{hisup} \& SAMPolyBuild~\citep{sampolybuild}.

% Interestingly, when viewing DragOSM as an SDE-based\footnote{Stochastic Differential Equation (SDE)} Wiener process~(\cref{sec:dragosm}), this improvement, from the 5-step denoising~\citep{li2025dragosm} in DragOSM to the 2-step efficient correction in DragRoof, mirrors the transition from DDPM~\citep{NEURIPS2020_ddpm} to flow-matching~\citep{lipman2022flowmatching} in image generation.

In conclusion, the main contributions of this work are:
\begin{itemize}
% 我们decouple了一个新任务：
    \item We advocate for decoupling RFOV extraction from segmentation and formally define it as a standalone, critical computer vision task. We analyse how existing coupled approaches hinder progress and argue that this new formulation is essential for current studies.
    \item We constructed ObliCity, the first large-scale benchmark dataset, including high-resolution UAV and global satellite imagery, for the challenging RFOV extraction. We will continue to update ObliCity to grow in size and scope for evolving real-world conditions. 
    \item  Inspired by the human annotating process, we propose a strong baseline method, DragRoof, with ODE modelling, which demonstrates the feasibility of the RFOV extraction problem. 
\end{itemize}

\section{Related Work}
In this section, we first review the evolution of building extraction methods from near-nadir to off-nadir imagery, and subsequently outline how our work bridges the existing gaps in this field.

\textbf{Near-nadir Imagery.} Research on building extraction initially focused heavily on near-nadir images, establishing a well-explored mainstream methodology. Prominent datasets, such as the WHU Building Dataset~\citep{ji2018whu-building}, SpaceNet~\citep{van2018spacenet}, and WHU-Mix~\citep{wei2023buildmapper}, have provided vast amounts of annotated building roofs. These datasets primarily consist of high-resolution remote sensing images captured from near-nadir perspectives, which successfully supported early-stage algorithm development~\citep{hisup, sampolybuild, vit_building_extraction, P2PFormerV2}. Consequently, the resulting methods are generally restricted to roof segmentation under ideal near-nadir assumptions. While the geometric projection displacement inherent in off-nadir imagery was widely acknowledged during this stage, it remained largely unsolved due to its complexity.

\textbf{Off-nadir Imagery.} Driven by the continuous advancement of deep learning, researchers have increasingly turned their attention to building extraction in off-nadir settings~\citep{BONAI, pang2023detecting, MTBRNet, MLS-BRN, obm, polyfootnet, zhou2025nesf}. This shift is motivated by the practical advantages of off-nadir acquisition: relaxed satellite viewing angle constraints significantly improve Earth observation efficiency, and the oblique perspective captures rich facade information essential for reconstructing LoD-2 level building models~\citep{groger2012ogc}, including building heights. However, current methodologies still rely on conventional semantic segmentation metrics (\eg, mask IoU) for evaluation, which fail to accurately reflect the actual quality of the extracted Roof-to-Footprint Offset Vectors (RFOVs) (detailed in Sec.~\ref{supp:rfov_mask}). Furthermore, existing off-nadir datasets, such as BONAI~\citep{BONAI} and OmniCity~\citep{li2023omnicity}, are constrained by the absence of 0.1m ultra-high-resolution UAV data, a lack of ultra-long RFOV samples, and relatively crude annotation boundaries.

To address these barriers, we formally decouple RFOV extraction into an independent computer vision task and introduce a comprehensive new dataset, ObliCity, to facilitate further research. Inspired by the natural human annotation process, we propose DragRoof as a robust baseline method tailored to effectively solve the RFOV extraction problem.

\section{Dataset: ObliCity}
\label{sec:dataset_creation}
In this section, we will first introduce why we newly annotated the ObliCity dataset

% 我们在数据集中融合
\subsection{Collection \& Annotation of ObliCity}
\textbf{Motivation.} ObliCity is proposed to address the limitations of existing datasets for RFOV learning, which are mainly captured from a single urban environment and rely solely on satellite imagery. We newly annotated UAV images (0.1\,m) to diversify the image resolution, and added RFOV annotations to the urban scenes of IRSAMap~\citep{meng2025irsamap} to include worldwide samples. 

\textbf{Image collection.}
% To enrich scene diversity, 我们对已有数据集IRSAMap数据集的建筑物数据进行了再次标注，进而获得每个建筑物的RFOV。IRSAMap是一个囊括了全球79个region的卫星数据集，覆盖了丰富的城市类型，共计有5,869张1024*1024的大尺度精细化的标注数据。我们对这批数据的城市oblique影像部分共666张进行了重新标注，分辨率0.5m。
% 同时，UAV数据集包括了长三角地区的7个不一样的城市地域，经统一标注和裁剪后，共1008张1536*1536的图片，分辨率约0.1m。For privacy and legal compliance, geolocation metadata has been anonymized. The dataset, including imagery and annotations, will be released upon publication.
% BONAI数据集则包括3300张1024*1024的0.3-0.5m的卫星遥感影像，共包含7个中国的大型城市，\eg, 成都，西安，哈尔滨等。
The images are collected from 3 parts: IRSAMap, BONAI~\citep{BONAI} and UAV. Specifically, to further enrich scene diversity, we re-annotate the building regions in the {IRSAMap~\citep{meng2025irsamap}} dataset with corresponding RFOVs. IRSAMap is a global satellite dataset covering 79 regions across various urban types, containing 5,869 finely annotated images of size 1024$\times$1024. Among them, we re-labeled 666 off-nadir images with a spatial resolution of 0.5\,m.
The {UAV dataset} includes images collected from seven distinct urban areas across the Yangtze River Delta region. After standardised annotation and cropping, it consists of 1,008 images of size 1536$\times$1536 with a spatial resolution of approximately 0.1\,m. 
% For privacy and legal compliance, all geolocation metadata has been anonymised. The dataset, including imagery and annotations, will be released upon publication.
The {BONAI dataset} contains 3,300 satellite images of size 1024$\times$1024 with a spatial resolution of 0.3–0.5\,m, covering six major cities in China, \eg, Chengdu, Xi’an, and Harbin.

% UAV影像的标注流程分为三步，针对未标注的影像，首先，我们人工地为每个建筑物绘制屋顶，然后，拖拽这些roof到其对应的建筑物footprint轮廓的可见部分重合。最后，基于拖动的轨迹，计算RFOV。IRSAMap由于已有了屋顶，因此只执行（2）（3），而针对BONAI存在半自动标注或不准确的部分，我们对数据集进行了再次的人工核验
\textbf{Annotation.}
The annotation process for UAV imagery consists of three steps. For each unannotated image, (1) annotators manually delineate the roof of every building, (2) drag the roof polygon until its edge aligns with the visible portion of the corresponding footprint, and (3) compute the RFOV based on the displacement trajectory of this drag operation.
For the {IRSAMap} dataset, we keep the roof annotations, and only steps (2) and (3) are performed. For {BONAI}, we keep it all. 
Inspired during this annotation process, we use an ODE-based modelling to design our DragRoof. 

To ensure the quality of dataset, each image was independently checked by another annotator.
Conflicts larger than 10 pixels were re-annotated.

\subsection{Properties of ObliCity}
%%%% 展示数据集图片
% ObliCity包括多样化的场景和卫星分辨率，并可以分为三部分，BONAI，IRSAMap, UAV。
% BONAI采样自中国快速发展的城市区域，建筑物之间空隙较大，且同范围内的建筑物外形风格类似，呈现出高度一致的RFOV。
% IRSAMap是一个全球化的数据集，其中包括世界各地的城市，受人文因素和百年的城市发展影响，影像的layout会比BONAI的城市更加复杂。
% UAV部分的数据则由于极高的影像分辨率，图像内囊括了非常细致的城市纹理和建筑物纹理，因而建筑物的屋顶标注更加复杂，这些建筑物的RFOV往往会比BONAI和IRSAMap中的更长，并且在图像切割时，必须保证较大的滑窗尺度（1536*1536），从而才能尽可能地囊括在切片中囊括完整的建筑物。UAV影像由于相机对地面位置更低，单次获得的影像区域更小，同时需要频繁的拼接才可以获得大尺度的影像，从而导致即使在同一个切片内，建筑物的RFOV方向也互相不同这与BONAI和IRSAMap中的情况完全不同。
The {ObliCity} dataset comprises diverse urban scenes and multiple spatial resolutions, consisting of three parts: {BONAI}, {IRSAMap}, and {UAV}. 
The {BONAI} subset is sampled from rapidly developing cities in China, while the {IRSAMap} subset provides global coverage, including cities across different continents. Influenced by diverse cultural and historical developments, its urban layouts are more complex and heterogeneous compared to those of BONAI.  

\begin{figure}
    \centering
    \includegraphics[width=\linewidth]{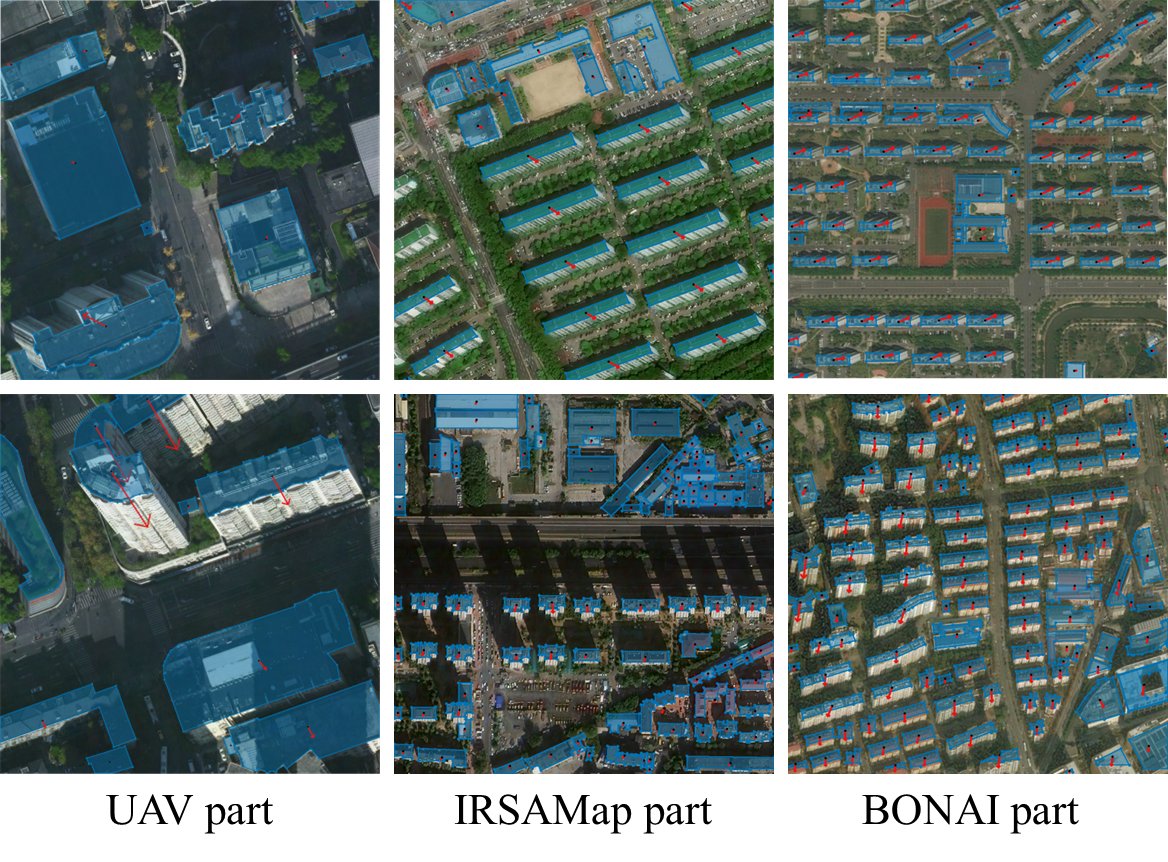}
    \caption{Samples in ObliCity. The ObliCity dataset is designed for the task of extracting RFOVs (red arrows) for buildings with known roofs. Each sample is created by manually delineating the roof and recording the vector obtained from dragging it to its corresponding footprint position.}
    \label{fig:dataset}

\end{figure}

In comparison, the {UAV} subset, captured at an ultra-high spatial resolution (0.1\,m), contains fine-grained urban and architectural textures. Its buildings often exhibit significantly longer RFOVs than those in BONAI and IRSAMap. During tiling, a larger sliding window (1536$\times$1536) is required to ensure complete building coverage. Besides, as UAV imagery is captured from lower altitudes with limited ground coverage per frame, frequent mosaicking is needed to obtain large-area views. Consequently, buildings within the same tile may have RFOVs pointing in different directions, a phenomenon rarely observed in BONAI or IRSAMap. This difference can be found in Fig.~\ref{fig:dataset}. % 插入图像2*3

\subsection{Dataset Splits}
% ObliCity被划分为训练集和测试集两部分。针对BONAI part，我们保留了其原有的划分，而IRSAMap则以600张训练集，66张作为测试集进行划分。UAV的部分则按照获取时的7个不同的区域，以5个整区域作为训练集，2个为测试集。Tab.~Table~\ref{tab:ObliCity_stats}中详细介绍了ObliCity数据集。
The {ObliCity} dataset is divided into training and testing subsets. For the {BONAI} part, we retain its original split. The {IRSAMap} subset is divided into 600 training images and 66 testing images. For the {UAV} subset, data are grouped by seven distinct acquisition regions, with five regions used for training and the remaining two for testing. Table~\ref{tab:ObliCity_stats} summarizes the detailed composition of the ObliCity dataset.

\begin{table}[t]
\centering
\caption{Statistics of the ObliCity dataset.}

\resizebox{\columnwidth}{!}{
\begin{tabular}{lrrrrrr}
\toprule
\multirow{2}{*}{{Data}} & \multirow{2}{*}{{Res. (m)}} & \multirow{2}{*}{{Size}} &
\multicolumn{2}{c}{{Train}} & \multicolumn{2}{c}{{Test}} \\
\cmidrule(lr){4-5} \cmidrule(lr){6-7}
 &  &  & {\# Img.} & {\# Ins.} & {\# Img.} & {\# Ins.} \\
\midrule
BONAI     & 0.3--0.5 & 1024$\times$1024 & 3,000 & 247,683 & 300 & 21,280 \\
IRSAMap   & 0.5       & 1024$\times$1024 & 600   & 52,244  & 66  & 3,893  \\
UAV       & 0.1       & 1536$\times$1536 & 827   & 16,599  & 181 & 4,936  \\
\midrule
{Total} &  &  & {4,427} & {316,526} &{547} & {30,109} \\
\bottomrule
\end{tabular}}
\label{tab:ObliCity_stats}

\end{table}

\section{Method: from DragOSM to DragRoof}

\subsection{Problem Setup}
Given an oblique remote sensing image $I$, the task of projection-difference extraction aims to derive the roof-to-footprint offset vector (RFOV) for each pre-located building instance. In other words, it serves as a post-detection task of building detection that provides a more precise description of structural characteristics, \eg, geometric configuration and relative height.

% In this work, we propose to solve this problem by supervised learning of an RFOV extraction model with dataset $\mathcal{D}$, \ie, $\mathcal{D} = \{ ({I}_i, \mathcal{\textbf{R}}_i, \mathcal{\textbf{O}}_i); i=1,..., N \}$, where $\mathcal{\textbf{R}}_i$ is pre-extracted building roofs in ${{I}}_i$, and $\mathcal{\textbf{O}}_i$ is the RFOVs. We use $\{\mathbf{r}_j, \Vec{o}_j; j=1,...,M\}$ to index each instance of $\mathcal{\textbf{R}}_i, \mathcal{\textbf{O}}_i$ in $\mathcal{\textbf{I}}_i$, and the target function of model is,
In this work, we propose to solve this problem by supervised learning of an RFOV extraction model with dataset $\mathcal{D}$, \ie, $\mathcal{D} = \{ ({I}_i, \mathbf{{R}}_i, \vec{ \mathbf{O}}_i); i=1,..., N \}$, where $\mathbf{R}_i$ is pre-extracted building roofs in ${{I}}_i$, and $\vec{ \mathbf{O}}_i$ is the RFOVs. We use $\{\mathbf{r}_j, \Vec{o}_j; j=1,...,M\}$ to index each instance of $\mathbf{R}_i, \vec{ \mathbf{O}}_i$ in ${I}_i$, and the target function of model is,
\begin{equation}
\label{eq:function}
    \Vec{o}_j=\mathrm{MODEL}({{I}}_i, \mathbf{r}_j).
\end{equation}
An example of this relationship is demonstrated in Fig.~\ref{fig:intro}(c). 

\begin{figure*}
    \centering
    \includegraphics[width=\linewidth]{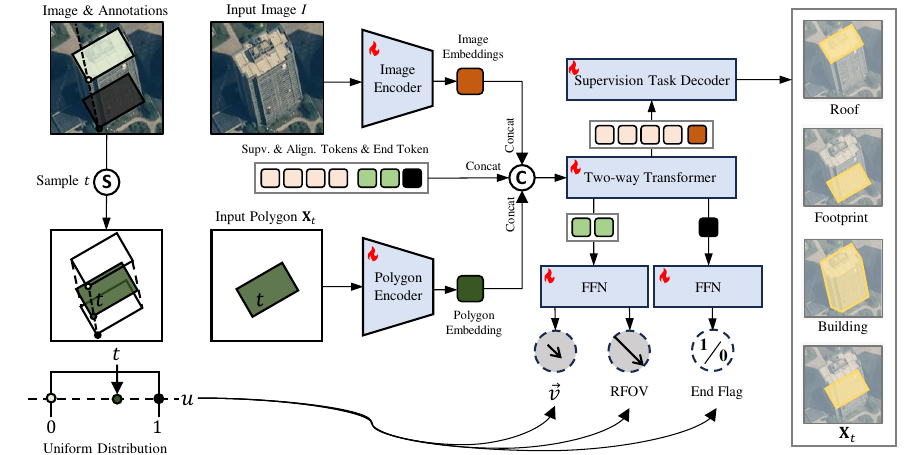}
    \caption{DragRoof in training. The system first samples a $t$ to get the input polygon; then this polygon and image will be encoded as embeddings. These embeddings, together with the tokens, will transfer their features through a Two-way Transformer, and finally, they will be decoded into masks, offset vectors, and an end flag, whose ground truth values are generated during the sampling process. DragRoof simulates the human annotation process by iteratively adjusting roof positions toward footprints with facade constraints through an ODE-based sampling scheme.}
    \label{fig:main_fig}

\end{figure*}

\subsection{Preliminary: DragOSM}
\label{sec:dragosm}
% DragOSM的提出是为了解决历史地图中的建筑物标签和新获取影像之间的不匹配问题。由于历史标签与影像中的建筑物存在随机的position discripancies. 为了模拟这种随机性，DragOSM在训练时通过给真值的标签叠加2D的高斯噪声来模拟真实世界的位置偏移，同时时训练模型来消除这种偏差。在推理的的过程中，标签的校正分为两步，首先通过一个连续的去噪过程，消除历史标签到建筑物footprint的位置偏差，然后再输入去噪的结果footprint，获得与图像语义一致的roof。这意味着，想要使用DragOSM提取RFOV就必须将起始位置的roof的输入理解为DragOSM输入的历史标签，该过程可以描述为SDE。
DragOSM was originally proposed to address the misalignment between historical building labels in maps and newly acquired imagery. Because these labels exhibit random positional discrepancies relative to the true building locations, DragOSM introduces 2D Gaussian noise to the ground-truth annotations during training to simulate such randomness, while jointly learning to correct it.  
During inference, the correction process is performed in two stages: (1) a continuous denoising procedure first eliminates the positional error between the historical label and the building footprint; and (2) the denoised footprint is then used to generate a roof polygon consistent with the image semantics.  
Therefore, applying DragOSM to RFOV extraction requires interpreting the initial roof input as the ``historical label'' in DragOSM’s formulation. This process can be mathematically described as an SDE\footnote{Stochastic Differential Equation (SDE)},
\begin{equation}
    d\mathbf{X}_t = \mathbf{f}(\mathbf{X}_t, t)\,dt + \sigma\,d\mathbf{W}_t,
    \label{eq:sde}
\end{equation}
where $\mathbf{X}_t \in \mathbb{R}^{2l}$ represents the concatenated coordinates of $l$ polygon vertices at time $t$, $\mathbf{W}_t$ is a Wiener process, and $\sigma$ controls the intensity of the positional noise. 
The term $\mathbf{f}(\mathbf{X}_t, t)$ denotes the \emph{alignment offset} generated by DragOSM. By discretizing Eq.~\ref{eq:sde} with step size $\Delta t$, we obtain the following iterative formulation:
\begin{equation}
    \mathbf{X}_{t+\Delta t} = 
    \mathbf{X}_t + \mathbf{f}(\mathbf{X}_t, t)\,\Delta t 
    + \sigma\sqrt{\Delta t}\,\boldsymbol{\epsilon},
    \quad \boldsymbol{\epsilon}\!\sim\!\mathcal{N}(\mathbf{0},\mathbf{I}).
    \label{eq:euler}
\end{equation}
Let $a_t$ absorb $\Delta t$ and other scaling coefficients, and denote 
$\mathbf{q}(\mathbf{X}_t;\theta,I)=\mathbf{f}(\mathbf{X}_t, t)\,\Delta t$. 
Then Eq.~(\ref{eq:euler}) simplifies to the discrete update rule:
\begin{equation}
    {\mathbf{X}_{t+1} = \mathbf{X}_t + a_t\,\mathbf{q}(\mathbf{X}_t;\theta,I)},
    \label{eq:update}
\end{equation}
which mirrors the multi-step denoising process in DragOSM~\citep{li2025dragosm} with model weight $\theta$, 
where $a_t=\delta^{t-1}$ is a geometrically decaying step size, and $\delta$ is a constant. 

During both training and inference, the relationship between the model-level \textit{alignment token} and the physical-level \textit{alignment offset vector} is formulated as:
\begin{equation}
\label{eq:align_decoder}
\vec{v}_e = \frac{\vec{v} - \vec{\nu}}{\kappa},
\end{equation}
where $\kappa$ is a scaling constant that normalises the physical vectors to a trainable range; $\vec{v}_e$ and $\vec{v}$ denote the encoded and decoded vectors, respectively; and $\vec{\nu}$ represents the mean alignment centre in the latent space.

\subsection{DragRoof}
\label{sec:dargosmp}

% Extracting RFOVs is different from updating the position of historical labels on new images. 因为历史标签在新影像上可能已经存在了偏移，这导致历史标签到footprint的校正只能以\cref{sec:dragosm}建模为SDE过程。但对于RFOV提取而言，以SDE建模该过程是不必要的，因为一旦图像确定，同时建筑物作为条件被指定 as Fig.~\ref{fig:main_fig}，RFOV提取变成为了一个确定性的过程，在图像上有着明确的语义特征。这种确定性破坏了DragOSM中的二维高斯假设，从而在RFOV提取任务上表现欠佳。
% We therefore reformulate the task as a deterministic ODE process to model RFOVs more accurately【\ie, only consider以footprint为原点，沿给定图像的偏移方向的切面方向为去噪方向，将不确定的2D高斯分布转化为了确定的1维分布 as Fig.~\ref{fig:main_fig}(c)】, and we name this as DragRoof. 
\textbf{Motivation.} Extracting RFOVs fundamentally differs from updating historical labels on new imagery, which requires an SDE model. Once the image and the corresponding building are fixed, the direction and magnitude of the RFOV become deterministic, directly governed by clear semantic and geometric cues in the image. 
% 这为我们设计模拟人工标注过程的ODE训练流程提供了可能, \ie, sampling the trajectory between roofs and footprints as training targets. This transformation converts the uncertain 2D Gaussian distribution into a deterministic 1D flow.
This deterministic nature naturally aligns with an ODE-based formulation, \ie, sampling the trajectory between roofs and footprints that simulates the human annotation procedure, and trains the model to predict the rest of the trajectory.

\textbf{Model structure. }
% DragRoof与DragOSM一样，有着几乎相同的模型结构。DragRoof也构建于SAM之上，使用ViT做为视觉编码器，以及一个多层的卷积网络用于编码输入的polygons。被编码的polygon feature将和两个DragOSM原有的两个alignment tokens同时输入twoway transformer的解码器，用于回归RFOV以及从当前polygon位置到footprint的offset vector $\Vec{v}_t$。同时，SAM中原有的mask预测功能将作为辅助任务，被用于回归polygon的当前输入位置，建筑物屋顶，底座，以及整个建筑物的分割，以帮助DragRoof更好地理解建筑物的结构。最后，为了让DragRoof自适应的感知输入的polygon位置是否已经到达了footprint的终点，DragRoof新增了一个end token，end token与$\Vec{v}_t$的长度高度绑定，当$\Vec{v}_t$的长度过小时，end token将被解码为1，反之为0.
As shown in Fig.~\ref{fig:main_fig}, {DragRoof} adopts an similar structure as DragOSM~\citep{li2025dragosm}. A Vision Transformer (ViT)~\citep{vit} is used as the visual encoder, while a multi-layer convolutional network encodes the input polygons $\mathbf{X}_t$. The encoded polygon features are concatenated with the two \textit{alignment tokens} from the original DragOSM and fed into a two-way Transformer decoder to regress both the RFOV and the offset vector $\Vec{v}_t$ from the current polygon $\mathbf{X}_t$ to the footprint. 
To allow the model to adaptively determine whether the current polygon has reached the footprint, we introduce an additional \textit{end token}. The activation of this token is closely coupled with the magnitude of $\Vec{v}_t$: it is decoded as 1 when $\|\Vec{v}_t\|$ becomes sufficiently small, indicating convergence, and 0 otherwise. 
Finally, the alignment tokens and end token will be decoded separately with different Feed-Forward Networks (FFNs).
Furthermore, the mask prediction branch of SAM is retained as an auxiliary task to segment the input polygon position, building roof, footprint, and overall building region, enabling DragRoof to better capture structural cues.  

\subsection{ODE Modelling}
\textbf{Motivation.} As introduced in Sec.~\ref{sec:dargosmp}, the core design of DragRoof is inspired by the human annotation process. 
In this section, we integrate this concept with the mathematical formulation of an ODE to model the training process.
% While the original DragOSM treats label correction as a stochastic denoising process formulated by an SDE, 
% the roof-to-footprint regression in DragRoof is inherently \emph{deterministic}: 
% given the roof geometry and the image $I$, there exists a unique displacement field describing the relative field of view (RFOV). 
% Therefore, we reformulate this task as a continuous and deterministic dynamic system governed by an ODE.

\textbf{Continuous formulation \& objective.}
Let $\mathbf{R}\!\in\!\mathbb{R}^{2l}$ denote the roof polygon (with $l$ vertices) and 
$\mathbf{F}\!\in\!\mathbb{R}^{2l}$ the corresponding footprint. 
We define a trajectory $\mathbf{X}_t$ that continuously moves the roof toward the footprint:
\begin{equation}
    \frac{d\mathbf{X}_t}{dt} = 
    \mathbf{v}_\theta(\mathbf{X}_t,t;I,\mathbf{R}), 
    \quad \mathbf{X}_0 = \mathbf{R}, \ 
    \mathbf{X}_1 = \mathbf{F},
    \label{eq:ode}
\end{equation}
where $\mathbf{v}_\theta$ is a learnable velocity field predicted by the Transformer decoder of DragRoof, conditioned on both the image and the current polygon state.
For the footprint location has a clear semantic feature among the given image, to accelerate the inference,
This field encodes the instantaneous motion that progressively aligns the roof with its ground footprint, \ie, $\mathbf{v}_\theta(\mathbf{X}_t,t;I,\mathbf{R}) = \Vec{v}_t$.
Because the displacement between roof and footprint is deterministic, we can sample the $\mathbf{X}_t$ by,
\begin{equation}
    \mathbf{X}_t = 
    \mathbf{R} + \phi(t)\,(\mathbf{F}-\mathbf{R}), 
    \quad \phi(0)=0,\ \phi(1)=1.
    \label{eq:sample}
\end{equation}
In practice, $\phi(t)$ can be a linear schedule ($\phi(t)=t$), and $t$ is sampled from a unified distribution $\mathcal{U}(0,1)$ to simulate the insufficient prediction of model, \ie, when $t=0$, it represents the initial state, footprint label on roof, and when $t=1$, it represents the roof label accurately placed on the footprint location; then, naturally the $\Vec{v}_t=\mathbf{F}-\mathbf{R}- \phi(t)\,(\mathbf{F}-\mathbf{R})$. 
The network is trained to match this target velocity via \emph{conditional flow matching}:
\begin{equation}
    \mathcal{L}_{\mathrm{CFM}}
    = \mathbb{E}_{t\sim\mathcal{U}(0,1)}
    \bigl\|
    \mathbf{v}_\theta(\mathbf{X}_t,t;I,\mathbf{R})
    - \Vec{v}_t
    \bigr\|_1.
    \label{eq:cfm}
\end{equation}

From Eq.~\ref{eq:sample}, each $\mathbf{X}_t$ will be semantically attached to the building selected in the given image $I$. Thus, a global supervision is added to regress the RFOV. As a result, the loss function is,
\begin{equation}
    \mathcal{L} = \alpha\mathcal{L}_{\mathrm{CFM}} +  \beta\mathcal{L}_{\mathrm{RFOV}}+ \gamma\mathcal{L}_{\mathrm{END}}+\sum_{n=1}^4 \lambda_n\mathcal{L}_n,
\end{equation}
where $\mathcal{L}_{\mathrm{RFOV}}$ denotes the smooth L1 Loss~\citep{fastrcnn} for the global RFOV supervision, $\mathcal{L}_{\mathrm{END}}$ denotes the Binary CrossEntropy Loss for end token flag, and $\sum_{n=1}^4\mathcal{L}_n$ is the CrossEntropy Loss~\citep{celoss} for the auxiliary mask tasks mentioned in \cref{sec:dargosmp}. In training, we use constants $\alpha, \beta, \gamma, \lambda$ to balance the losses. 

\textbf{Training.}
As a result, the training pipeline is as Fig.~\ref{fig:main_fig}: given an image and a building roof annotation, the model will first sample a $t$ to get an interval polygon. From here, the ground truth of $\Vec{v}$, RFOV and End Flag, used for supervised learning. 

Then, the image and polygon will be encoded as embeddings. They will consequently be concatenated with supervise mask tokens, align tokens and end tokens. The following Two-way Transformer will weave their features. 
Finally, these tokens will be decoded as masks, $\Vec{v}$, RFOV and end flag. The $\Vec{v}$ and RFOV are encoded in the same way as Eq.~\ref{eq:align_decoder} in DragOSM. 

\textbf{Inference.}
At inference, the model integrates Eq.~(\ref{eq:ode}) numerically using a $K$-step iterative system to get the predictions $\{\hat{\mathrm{RFOV}_k}\}_{k=1}^K$, end flag $\{\hat{e}_k\}_{k=1}^K$, and the predicted $\{\hat{\Vec{v}}_k\}_{k=1}^K$:
\begin{align}
    \mathbf{X}_{k+1} &= 
    \mathbf{X}_k +    \mathbf{v}_\theta(\mathbf{X}_k,t_k;I,\mathbf{R}). 
    % \mathrm{RFOV} &= \sum_{k=1}^K \mathbf{v}_\theta(\mathbf{X}_k,t_k;I,\mathbf{R}).
    % \mathrm{RFOV}_k, \hat{e}_k,  &= 
    \label{eq:euler}
\end{align}
Starting from $\mathbf{X}_0=\mathbf{R}$, integration proceeds until $K$ steps. 
Finally, the global prediction $\hat{\mathrm{RFOV}}$ is defined as:
\begin{equation}
    \hat{\mathrm{RFOV}} = \sum_{k=1}^K \frac{g_1(\hat{e}_k)\hat{\mathrm{RFOV}}_k + g_2(\hat{e}_k)\hat{\Vec{v}}_k}{\sum_{k=1}^K g_1(\hat{e}_k) +1},
\end{equation}
where $g_1(\cdot)$ and $g_2(\cdot)$ are selection strategies based on the value of $\hat{e}_k$, and their outputs are 0 or 1.
In practice, we simply set both $g_1(\hat{e}_k)$ and $g_2(\hat{e}_k)$ after $\hat{e}_k>\eta$ as 0, otherwise 1, which means the end of the prediction. Specifically, if the model converges within a single inference step, $\mathrm{RFOV}_1$ is taken as the final output.

\section{Experiment}

\begin{table*}[t]
  \centering
  \setlength{\tabcolsep}{2pt}
  \scriptsize
  \caption{Main results on ObliCity test set and its test sub-sets.}

  \label{tab:rfov_mainresults}
  \resizebox{\linewidth}{!}{%
  \begin{tabular}{lr|cccccccccccccccc}
    \toprule
    \multirow{2}{*}{Model} & \multirow{2}{*}{Type}
      & \multicolumn{11}{c}{E$_i$$\downarrow$} & \multirow{2}{*}{mE$\downarrow$}
      & \multicolumn{4}{c}{aE$\downarrow$} \\
    \cmidrule(lr){3-13}\cmidrule(l){15-18}
     &  & 10 & 20 & 30 & 40 & 50 & 60 & 70 & 80 & 90 & 100 & $\infty$
        &  & ObliCity & BONAI & UAV & IRSAMap \\
    \midrule
    \multirow{3}{*}{LOFT~\citep{BONAI}}
      & EPE & 6.91 & 7.30 & 10.01 & 15.86 & 19.23 & 24.14 & \textbf{28.34} & \textbf{38.24} & 34.83 & 39.24 & 142.77 & 33.35 & 10.61 & 8.21 & 26.03 & 4.56 \\
      & LE & 4.81 & 4.51 & \textbf{6.60} & 10.59 & 13.49 & \textbf{18.23} & \textbf{21.40} & \textbf{30.90} & 28.22 & 26.85 & 128.63 & 26.75 & 7.89 & 5.91 & 20.03 & \textbf{3.48} \\
      & AE & 0.80 & 0.38 & 0.32 & 0.38 & 0.38 & 0.39 & 0.36 & 0.44 & 0.33 & 0.41 & 0.70 & 0.45 & 0.49 & 0.44 & 0.68 & 0.50 \\
    \midrule
    \multirow{3}{*}{PolyFootNet~\citep{polyfootnet}}
      & EPE & 6.44 & 7.19 & 9.53 & 13.89 & 21.82 & 28.21 & 32.31 & 42.57 & \textbf{32.37} & \textbf{36.80} & 142.28 & 33.95 & 10.60 & 8.13 & 26.39 & \textbf{4.27} \\
      & LE & 5.30 & 5.46 & 7.02 & 10.84 & 16.91 & 23.36 & 27.56 & 36.93 & \textbf{27.27} & 29.97 & 132.64 & 29.39 & 8.86 & 6.78 & 22.15 & 3.57 \\
      & AE & 0.56 & 0.30 & 0.29 & 0.31 & 0.39 & 0.40 & 0.35 & 0.43 & 0.33 & 0.33 & 0.65 & 0.39 & 0.37 & 0.30 & 0.61 & 0.40 \\
    \midrule
    \multirow{3}{*}{DragOSM~\citep{li2025dragosm}}
      & EPE & 5.92 & 8.86 & 13.33 & 20.24 & 28.23 & 39.36 & 49.91 & 61.55 & 67.88 & 75.00 & 118.42 & 44.43 & 11.00 & 10.96 & 17.61 & 5.97 \\
      & LE & \textbf{3.63} & 6.45 & 10.74 & 16.70 & 24.03 & 35.91 & 46.74 & 57.24 & 61.39 & 69.95 & 111.03 & 40.35 & 8.72 & 8.67 & 14.41 & 4.43 \\
      & AE & 0.88 & 0.48 & 0.38 & 0.43 & 0.42 & 0.43 & 0.45 & 0.56 & 0.54 & 0.52 & 0.52 & 0.51 & 0.57 & 0.56 & 0.55 & 0.62 \\
    \midrule
    \multirow{3}{*}{Ours}
      & EPE & \textbf{5.01} &\textbf{ 5.95} & \textbf{9.14} & \textbf{11.96} & \textbf{15.71} & \textbf{22.22} & 30.53 & 44.11 & 38.98 & 44.17 & \textbf{89.01} & \textbf{28.80} & \textbf{7.99} & \textbf{7.67} & \textbf{14.08} & {4.59} \\
      & LE & 3.76 & \textbf{4.47} & 7.31 & \textbf{9.86} & \textbf{12.87} & 19.94 & 28.59 & 40.42 & 36.20 & 40.19 & \textbf{83.61} & \textbf{26.11} & \textbf{6.61} & \textbf{6.36} & \textbf{11.60} & {3.73} \\
      & AE & \textbf{0.53} & \textbf{0.25} & \textbf{0.21} & \textbf{0.19} & \textbf{0.17} & \textbf{0.16} & \textbf{0.21} & \textbf{0.29} & \textbf{0.18} & \textbf{0.17} & \textbf{0.21} & \textbf{0.24} & \textbf{0.32} & \textbf{0.30} & \textbf{0.38} & \textbf{0.37} \\
    \bottomrule
  \end{tabular}}
  \begin{tablenotes}[flushleft]  % ← 注释区，可选 [flushleft/right/center] 对齐
      \footnotesize
\item \textbf{Bold} highlights the best; 
$\downarrow$: lower is better; E$_i$, mE, and aE represent the range, mean, and average metrics of EPE, LE, and AE, respectively. 
For DragOSM, we evaluate all denoising configurations and report the best results under the optimal hyper-parameter setting  (1-step denoising).
    \end{tablenotes}
\end{table*}

\subsection{Implementation Detail}
We use ViT-Base~\citep{vit} initialized from DragOSM~\citep{li2025dragosm} as the backbone. Both DragOSM and DragRoof are fully fine-tuned on our new dataset. All experiments are conducted on a single server equipped with seven NVIDIA RTX GPUs (24\,GB each). We employ SGD with a weight decay of 0.0001 and a momentum of 0.9. The training runs for 48 epochs, with the learning rate warming up from 0 to 0.0025 over the first 500 iterations and decaying by a factor of 0.1 at the 32nd and 44th epochs. The $\eta$ for $g_1(\cdot)$ and $g_2(\cdot)$ is set to 0.5. % \footnote{The detailed \textit{Compute Reporting Form} and \textit{Training Logs} have been uploaded. Computational complexity is also included in \cref{supp:CC}.}
Our model contains 90.1M trainable parameters.
During inference, the model achieves 0.85 FPS on the full test set ($\approx$ 1.17 seconds per image) using a single NVIDIA RTX 3090 GPU. 
\subsection{Evaluation Protocol}
The evaluation of the RFOV extraction focuses on the quality of the predicted offset vectors. We assess performance from three complementary aspects between the predicted and ground-truth vectors: vector Angular Error (AE), vector Length Error (LE), and End-Point Error (EPE). For each prediction vector $\Vec{o}_p$ and the ground-truth vector $\Vec{o}_g$, the AE (in radian) is defined as the smaller angle between them, \ie, the one less than $180^\circ$, while AE and LE (in pixels) are measured as: 
\begin{align}
\mathrm{LE} &= \big|\big|||\vec{o_g}||_2 - ||{\vec{o_p}}||_2\big|\big|_2, \\
\mathrm{EPE} &= \left|\left|\vec{o_g} - {\vec{o_p}}\right|\right|_2,
\end{align}
where $||\cdot||_2$ represents the 2nd Norm. 
Based on these three metrics, for a test set, we group the vectors based on the length of the ground truth. EPE as an example, we define EPE$_i$ where $i=10, 20, ..., 100$ to represent the average EPE of predicted vectors whose length of ground truth vector is between $[i-10, i)$; differently, EPE$_\infty$ is used for the length interval $[100,\infty)$. Then, mEPE is defined as the mean of all EPE$_i$, and we use aEPE to denote the average EPE of all predictions in the test set. Finally, LE$_i$, AE$_i$, mLE, mAE, aLE, and aAE are defined in the same way.

\subsection{Revisions on BONAI}
\begin{figure}[t]
    \centering
    \includegraphics[width=\linewidth]{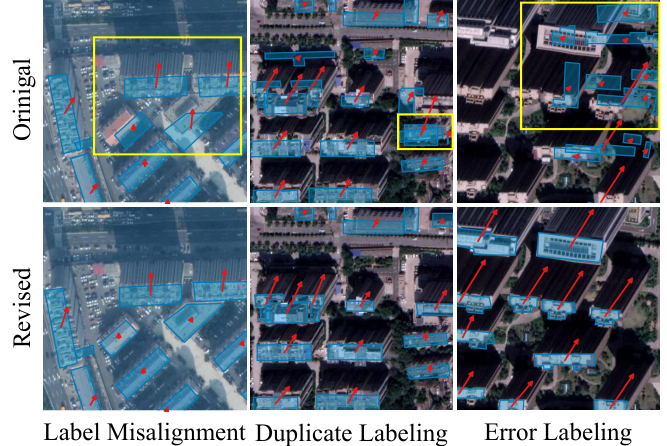}
    \caption{Examples of label revisions on the BONAI dataset. The selected cases show three common correction types: label-image misalignment, duplicate labeling, and error annotations. These revisions improve the consistency among roof masks, footprint masks, RFOVs, and the corresponding image content.}
    \label{fig:bonai_revision}
\end{figure}

Although the BONAI dataset includes roof masks, footprint masks, and RFOVs for most of the buildings, the labels suffer from problems, \eg, missing, mistaken labeling, and label-image misalignments. We use the tool of the Sec.~\ref{sec:dataset_creation} to revise the labels of BONAI to improve the quality of the training set. Fig.~\ref{fig:bonai_revision} illustrates representative revision examples, including newly added, deleted, and modified labels. Based on statistics, 634 labels are newly added, 1,320 labels are deleted, and 11,912 labels are modified. 

\subsection{Comparisons}

DragRoof is tailored for RFOV extraction. The function of the model needs to follow the Eq.~\ref{eq:function}. In this paper, we pick methods with a similar function: 
(1) \textbf{LOFT~\citep{BONAI}} is a two-stage model based on Mask RCNN~\citep{maskrcnn}. It supports manually providing a roof region of interest (roof bounding box), and uses its feature-level offset augmentation module to extract the related RFOV. (2) \textbf{PolyFootNet~\citep{polyfootnet}} is built upon OBM~\citep{obm}. It uses a mix-of-expert (MoE) system, named reference offset augmentation module, to predict RFOV in one step with given roof bounding boxes, and applies self-offset attention to improve the angle of RFOV. (3) \textbf{DragOSM~\citep{li2025dragosm}} is trained for map update under SDE assumption. It receives historical labels as prompts to predict two offsets to get roofs and footprints on updated images. 

LOFT and PolyFootNet were originally designed to assist the manual annotation of building footprint masks, and therefore use boxes as inputs to simplify human interactions. 
In contrast, DragOSM and DragRoof are mask-based label correction models that take roof masks as input.

\subsection{Results \& Analyses}
In Table~\ref{tab:rfov_mainresults}, we evaluate the RFOV extraction performance of our proposed method on ObliCity. The results are reported separately: overall results and sub-set results.

% LOFT通过解码裁剪的image feature map的roof部分的feature，获得RFOV，这种特征在空间上不足以覆盖一个建筑物的全部像素，因而模型虽然在几个少有的指标上有领先优势，但在数据集层面的指标表现最差。
% 相比之下，PolyFootNet是一个基于SAM的架构，细化了RFOV的提取，为不同长度的RFOV预测，设计了一个类似MoE的ROAM解码器作为解码预测RFOV，以及SOFA模块做全局的偏移量优化，因而有着几乎排名第二的表现。
\textbf{Overall results.} LOFT~\citep{BONAI} predicts RFOVs by decoding the roof features cropped from the image feature map. 
However, these features are spatially limited and fail to cover the entire building region. 
As a result, although LOFT achieves slight advantages on a few specific metrics, its overall performance on the dataset remains the lowest. PolyFootNet~\citep{polyfootnet} adopts a SAM-based architecture and refines the process of RFOV extraction. 
It employs a ROAM decoder, similar to a mixture-of-experts design, to handle RFOVs of varying lengths, and incorporates a SOFA module for global offset optimization. 
Consequently, PolyFootNet consistently ranks second across most evaluation metrics.
% DragOSM使用roof mask作为输入，本身相比于roof box有着更加明确的空间语义提示。然而，由于DragOSM是基于SDE的2D高斯假设训练，模型极可能为输入的屋顶匹配到错误的建筑物底座，因而只有在第一步去噪时的global offset可用于表示RFOV，因此仅在第一步去噪时，模型便达到了上限表现。但相比之下，采用随机分布位置的Polygon作为输的DragOSM相比于LOFT和PolyFootNet使用固定位置的roof box输入，在超长RFPV的预测上（E$_\infty$ & UAV sub-set）有着更加优秀的表现。
DragOSM~\citep{li2025dragosm} takes roof masks as input, providing a more explicit spatial prior than the roof boxes used by other methods. 
However, since DragOSM is trained under an SDE-based 2D Gaussian assumption, the model may associate a given roof with an incorrect building footprint. 
As a result, only the global offset estimated during the first denoising step can effectively represent the RFOV, and the model reaches its upper-bound performance at this stage (refer to Sec.~\ref{dis: SDE_ODE} and supplemental materials Sec.~\ref{supp:ode_sde}).  
Nevertheless, compared with LOFT and PolyFootNet, which rely on fixed roof boxes, DragOSM uses randomly distributed polygon inputs, leading to better performance in predicting ultra-long RFOVs (the E$_\infty$ and the aE on UAV subset), although only its first step estimation is valuable. 
% We will further probe this phenomenon in \cref{dis: SDE_ODE}.

%In comparison, DragRoof利用给定图像和建筑prompt时，建筑物的偏移方向作为先验条件已经暗含于图像中，因而设定了ODE的假设进行去噪训练，在感知RFOV的方向性上有着超乎前方所有方法的方向准确性。同时，去噪思想的融入，让DragRoof在几乎所有长度的RFOV预测上有着最优表现，\eg, aEPE在整个ObliCity上，相比于第二名的PolyFootNet低2.61pixel.
In comparison, DragRoof leverages the fact that, given an image and a building prompt, the directional prior of the building offset is already implicitly encoded within the image. 
Accordingly, it adopts an ODE-based formulation for denoising, which enables the highest accuracy of predicting RFOV directions (better at all AEs). 
Moreover, the iterative inference ability allows DragRoof to achieve superior performance across almost all RFOV lengths. 
\eg, its average EPE on the entire ObliCity dataset is 2.62 pixels lower than that of the second-best method, PolyFootNet.

\textbf{Subset results. }
% 通过跨数据集比较，UAV子集有着最大的aE，而在IRSAMap上最小。结合数据集本身的特点：UAV数据集以无人机等航空平台在地空获取，有着极高的分辨率，因此RFOV长度大（通常超过100pixel最长甚至接近1000 pixels）；而BONAI和IRSAMap的主要差别则体现在数据集的采样地点，BONAI主要采样了中国境内的高楼林立的大城市，IRSAMap作为一个全球采样的数据集，RFOV extraction的任务难度可能并不与建筑物风格直接相关，而是与RFOV本身的长度相关的任务, 当RFOV越长，难度越大，RFOV越短，难度越小。
Across datasets, the UAV subset exhibits the highest aE, while IRSAMap shows the lowest, \eg, aEPE of UAV predicted by DragRoof nearly doubled that on BONAI. 
This observation aligns with the characteristics of each dataset: the UAV subset is captured from aerial platforms, \eg, drones, offering ultra-high spatial resolution. 
Consequently, the RFOVs in UAV imagery are much longer, \ie, exceeding 100 pixels and sometimes approaching 1,000 pixels. 
In contrast, BONAI and IRSAMap mainly differ in sampling regions. BONAI focuses on densely populated metropolitan areas in China, whereas IRSAMap covers a diverse range of global cities. 
Overall, the difficulty of RFOV extraction appears to correlate more strongly with the length of the RFOV than with the architectural style: longer RFOVs are inherently more challenging to predict, while shorter ones are relatively easier.

\begin{figure*}
    \centering
    \includegraphics[width=0.95\linewidth]{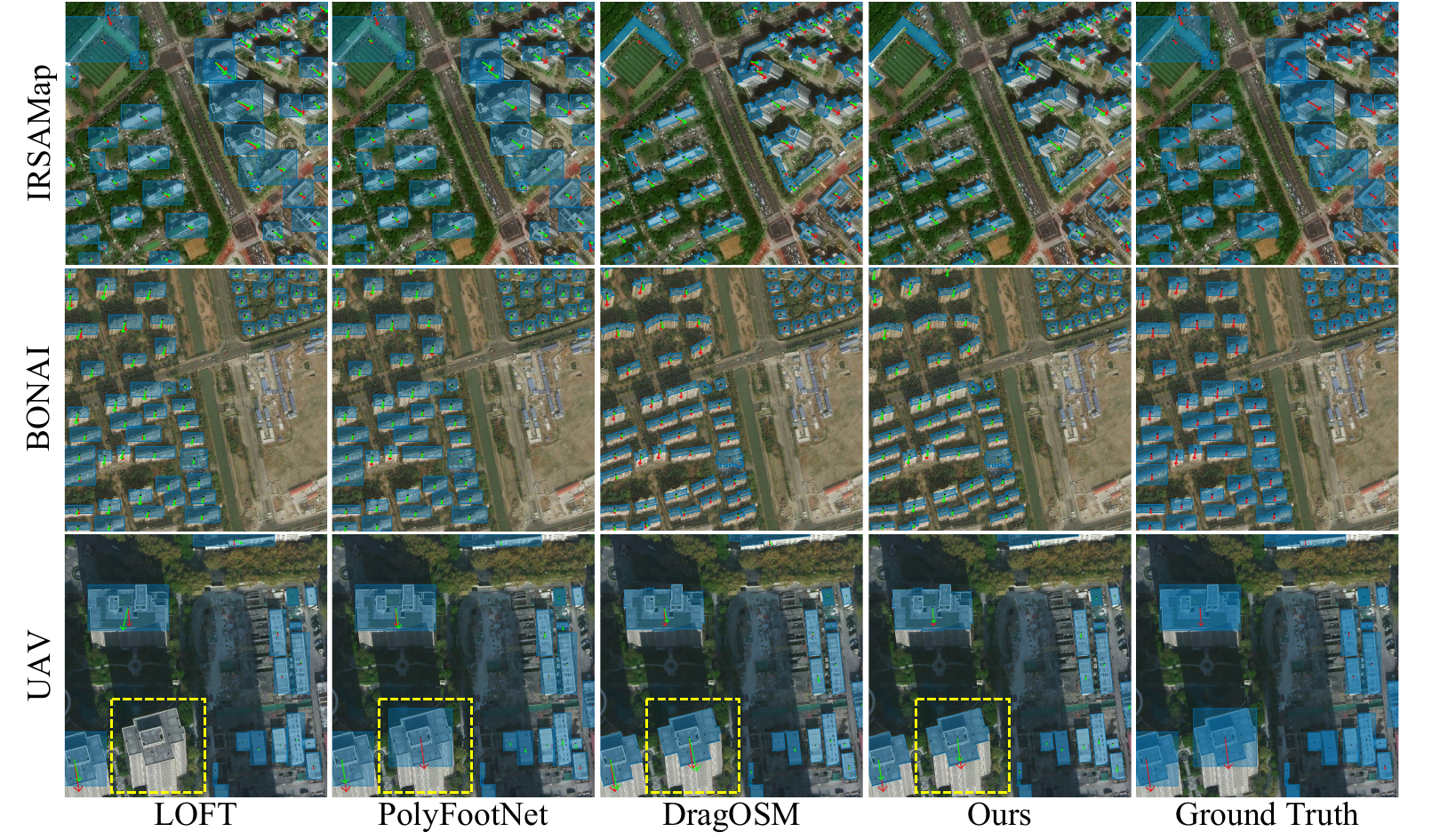}
    \caption{Visualized results on three parts of the ObliCity test set. 
Each example shows the input prompts (blue bounding boxes or polygons) and the model predictions (green arrows) overlaid on the ground-truth RFOVs (red arrows) for easier comparison across methods.
}
    \label{fig:test_results}
\end{figure*}

\textbf{Visualization \& case study\footnote{More visualizations \& generalization test in supplemental materials}. }
% 为了方便对结果进行理解，我们在Fig.1对结果进行了可视化。
To enhance the understanding of the mentioned results, we visualised them in Fig.~\ref{fig:test_results}. 
% 整体而言，RFOV提取任务的难度主要集中于ultra-long offset的预测。而我们的模型相比于 LOFT, PolyFootNet和DragOSM能够预测与ground truth RFOV更加一致的方向和长度（Fig.~\ref{fig:test_results}中预测结果与真值几乎重合），而mask prompt 相比于 box prompt对于物体的指示有更大的明确性。同时，过往的RFOV常作为实例分割类任务的辅助任务，这种耦合性严重影响了RFOV的提取。\eg，以黄色框区域failure case的建筑物为例。LOFT虽然被直接提供了真值的roof box作为输入，但由于其本身的classification head将输入的建筑物划分为了负样本，因而预测出现了缺失。而PolyFootNet则直接错误预测了RFOV. DragOSM的输出则存在较大的角度偏差。而我们的模型虽然预测出了极其一致的方向，但可能由于人工设置的Ending Flag使得去噪过程过早结束，从而导致长度过短。
Overall, the main challenge in RFOV extraction lies in predicting ultra-long offsets. 
As shown in Fig.~\ref{fig:test_results}, our model produces RFOVs that are more consistent with the ground truth in both direction and length, with many predictions nearly overlapping the annotations. 
The mask prompt offers a more explicit and spatially precise guidance than the box prompt, enabling more accurate feature localization.  
In previous works, RFOV prediction was often treated as an auxiliary task coupled with instance segmentation, which limited its independent optimization. 
\eg, in the failure case highlighted by the yellow box: LOFT, despite being provided with the ground-truth roof box, it misclassified the building as a negative sample, resulting in missing predictions. 
PolyFootNet, on the other hand, produced an incorrect RFOV, while DragOSM exhibited large directional deviations. 
In contrast, our model yields an almost perfectly aligned direction; however, the manually defined end flag threshold occasionally terminates inference too early, leading to slightly shorter RFOV lengths.

\begin{table}[t]
  \centering
  \setlength{\tabcolsep}{3pt}
  % \scriptsize
  \small
  \caption{Performance under varying flag thresholds on ObliCity.}

  \label{tab:flag_threshold_performance}
  \begin{tabular}{lcccccc}
    \toprule
    Flag Threshold & 0.5 & 0.6 & 0.7 & 0.8 & 0.9 & 0.95 \\
    \midrule
    aEPE & {7.99} & 7.97 & 8.00 & 8.03 & 8.10 & 8.18 \\
    aLE  & {6.59} & 6.59 & 6.62 & 6.66 & 6.73 & 6.83 \\
    aAE  & 0.323 & 0.322 & 0.320 & 0.321 & 0.322 & {0.322} \\
    \bottomrule
  \end{tabular}

\end{table}

\begin{table}[t]
  \centering
  \setlength{\tabcolsep}{3pt}
  % \scriptsize
  \small
  \caption{The stop rate (\%) in different flag thresholds over steps.}

  \label{tab:flag_stop_step}
  \begin{tabular}{ccccccc}
    \toprule
    \multirow{2}{*}{Step} & \multicolumn{6}{c}{Flag Threshold} \\
    \cmidrule(lr){2-7}
     & 0.5 & 0.6 & 0.7 & 0.8 & 0.9 & 0.95 \\
    \midrule
    1 & 96.80 & 94.72 & 91.30 & 85.28 & 70.39 & 48.95 \\
    2 & 98.82 & 97.79 & 95.94 & 92.57 & 82.46 & 62.71 \\
    3 & 99.42 & 98.72 & 97.58 & 95.40 & 88.32 & 72.52 \\
    4 & {99.70} & {99.20} & {98.33} & {96.69} & {91.02} & {77.29} \\
    \bottomrule
  \end{tabular}

\end{table}

% 与DragOSM的SDE假设不同，DragOSM的SDE训练将footprint的位置作为高斯分布的0均值中心，因而在连续校正是具备稳定性，\ie, 随着迭代的增多，输出的$X_t$的移动方向具有中心性，因而即使当X_t被错误的分配了标签位置，依据于2D高斯假设，对于OSM的输入仍然可以匹配一个高斯的0均值中心作为去噪的目标位置，进而在迭代的过程中具备稳定性。而DragRoof的ODE假设则基于均匀分布采样，X_t先验地分布在图像上building所蕴含像素的范围内，这意味着当输入的X_t位置不在图像上的建筑物范围内时，模型由于并没有学习到该内容，此时的输出完全可能会污染已有的去噪结果，因此我们需要一个End Flag让模型自主地判断是否需要停止预测。End Flag的加入即避免了致命的越界预测，同时自主地判断预测结果而避免了过度推理，带来了计算的高效性。
\textbf{Ablations.} Unlike the SDE assumption in DragOSM, where the footprint position is modelled as a Gaussian distribution centred at zero mean, the iterative correction process exhibits inherent stability. 
Specifically, as the number of iterations increases, the predicted displacement $\textbf{X}_t$ tends to move toward the distribution centre. 
Even when $\textbf{X}_t$ is initially assigned to an incorrect label position, the model can still converge toward the Gaussian mean under the 2D Gaussian assumption, enabling stable denoising for OSM-based inputs.
In contrast, DragRoof adopts an ODE formulation with uniformly sampled priors, where $\textbf{X}_t$ is distributed within the pixel range corresponding to the building area in the image. 
This means that if $\textbf{X}_t$ falls outside the valid building region, the model, having never learned such cases, may produce unpredictable outputs that corrupt the ongoing inference process. 
To address this, we introduce an \textit{End Flag} that enables the model to autonomously determine when to stop making predictions. 

To verify our observation, we set different end-flag thresholds to control when DragRoof stops inference, \ie, higher thresholds lead to later stopping steps. 
As shown in Table~\ref{tab:flag_threshold_performance}, a slight increase in the threshold initially improves performance; however, further increases gradually degrade it. 
Table~\ref{tab:flag_threshold_performance} also reports the average number of steps before the model stops predicting under different thresholds, showing that DragRoof typically converges within two steps. 
In contrast, DragOSM~\citep{li2025dragosm} requires about five denoising steps to align the historical map with the updated image.
As a result, combining the results in Table~\ref{tab:rfov_mainresults}, where DragRoof outperforms DragOSM, this finding indicates that DragRoof achieves superior performance with fewer inference steps. 
This behaviour aligns well with observations in image generation tasks, where ODE-based flow matching~\citep{lipman2022flowmatching} often attains comparable or better results than SDE-based DDPMs~\citep{NEURIPS2020_ddpm} using significantly fewer iterations.

% On the other hand, PolyFootNet与DragRoof都是基于SAM实现的，但是在提取RFOV时的方式不同，PolyFootNet使用了一个类似MoE的多头FFN（4个）解码架构，先前向推理一个基础输出再分别由专家头推理精确长度; 而DragRoof则只需要1个FFN，通过2步的连续去噪就获得了比PolyFootNet更加准确的RFOV，\ie, 仅使用了1/4的参数，和同样的推理步数就获得了更低的mEPE （\downarrow 5.15 pixels）.
On the other hand, both PolyFootNet and DragRoof are implemented upon SAM~\citep{sam}, but they differ in how RFOVs are extracted. 
PolyFootNet adopts a 4-head FFN decoder, resembling a MoE design. 
It first performs a forward inference to generate a coarse output and then refines the RFOV length through expert-specific predictions. 
In contrast, DragRoof requires only a single FFN and achieves more accurate RFOV estimation through two consecutive denoising steps, \ie, using only 1/4 parameters and the same number of inference steps, DragRoof attains a lower mean EPE ({$\downarrow$ 5.15 pixels}) than PolyFootNet (Table~\ref{tab:rfov_mainresults}).

\textbf{Impact of roof robustness.} 
Our experiments are conducted under the setting that reasonably accurate roof extraction is available. However, this assumption may not always hold in practical scenarios, since roof masks are typically obtained by external algorithms. Therefore, we add random noise to the ground-truth annotations of the ObliCity dataset to simulate varying roof-extraction quality (IoU) and analyze its impact on RFOV estimation.

% \begin{table*}
% \scriptsize
% \setlength{\tabcolsep}{2.5pt}
% \renewcommand{\arraystretch}{0.8}

% \centering
% \begin{minipage}{\linewidth}
% \centering
% \caption{DragRoof with noised roof inputs.}
% \label{tab:roof_effect}
% \begin{tabular}{lrrrrr}
% \toprule
% \multirow{2}{*}{IoU} & \multirow{2}{*}{mEPE} & \multicolumn{4}{c}{aEPE} \\
% \cmidrule(lr){3-6}
%  &  & Obli. & BON. & UAV & IRS. \\
% \midrule
% 1.00 & 28.80 & 7.99 & 7.67 & 14.08 & 4.59 \\
% 0.88 & 29.26 & 8.09 & 7.82 & 14.02 & 4.57 \\
% 0.65 & 29.36 & 8.16 & 7.89 & 14.02 & 4.71 \\
% 0.32 & 30.80 & 8.53 & 8.20 & 14.23 & 5.45 \\
% 0.18 & 33.18 & 9.04 & 8.71 & 14.81 & 5.92 \\
% \bottomrule
% \end{tabular}
% \end{minipage}
% \hfill
% \begin{minipage}{\linewidth}
% \centering
% \caption{Decoupled vs. non-decoupled extraction.}
% \label{tab:decp}
% \begin{tabular}{lrrrrrr}
% \toprule
% \multirow{2}{*}{Model} & \multicolumn{3}{c}{Roof} & \multicolumn{3}{c}{Footprint} \\
% \cmidrule(lr){2-4}\cmidrule(lr){5-7}
%  & Prec. & Rec. & F1 & Prec. & Rec. & F1 \\
% \midrule
% LOFT & 46.5 & 91.5 & 59.1 & 42.8 & 90.8 & 55.7 \\
% MLS.  & 56.5 & 86.1 & 66.2 & 53.9 & 84.6 & 63.8 \\
% Ours & 69.4 & 76.8 & 71.9 & 64.6 & 73.4 & 67.7 \\
% \bottomrule
% \end{tabular}
% \end{minipage}
% \end{table*}

\begin{table}

\centering
\caption{DragRoof with noised roof inputs.}
\label{tab:roof_effect}

\begin{tabular}{lrrrrr}
\toprule
\multirow{2}{*}{IoU} & \multirow{2}{*}{mEPE} & \multicolumn{4}{c}{aEPE} \\
\cmidrule(lr){3-6}
 &  & Obli. & BON. & UAV & IRS. \\
\midrule
1.00 & 28.80 & 7.99 & 7.67 & 14.08 & 4.59 \\
0.88 & 29.26 & 8.09 & 7.82 & 14.02 & 4.57 \\
0.65 & 29.36 & 8.16 & 7.89 & 14.02 & 4.71 \\
0.32 & 30.80 & 8.53 & 8.20 & 14.23 & 5.45 \\
0.18 & 33.18 & 9.04 & 8.71 & 14.81 & 5.92 \\
\bottomrule
\end{tabular}

\end{table}

%%%%%%%%%%%%%%%%%%%%%%
\begin{table}
\centering
\caption{Decoupled vs. non-decoupled extraction.}
\label{tab:decp}
\begin{tabular}{lrrrrrr}
\toprule
\multirow{2}{*}{Model} & \multicolumn{3}{c}{Roof} & \multicolumn{3}{c}{Footprint} \\
\cmidrule(lr){2-4}\cmidrule(lr){5-7}
 & Prec. & Rec. & F1 & Prec. & Rec. & F1 \\
\midrule
LOFT & 46.5 & 91.5 & 59.1 & 42.8 & 90.8 & 55.7 \\
MLS.  & 56.5 & 86.1 & 66.2 & 53.9 & 84.6 & 63.8 \\
Ours & 69.4 & 76.8 & 71.9 & 64.6 & 73.4 & 67.7 \\
\bottomrule
\end{tabular}

\end{table}

As Tab.~\ref{tab:roof_effect}, as the edge noise of the input roof masks increases, the IoU between the input roofs and the ground truth decreases, and the RFOV extraction error increases accordingly. However, the EPE degradation is relatively mild compared to the significant drop in roof quality. \eg, when the IoU drops to 0.65, the aEPE on ObliCity increases by only 0.17 ($\uparrow2.1\%$).
This suggests that, under the decoupled setting, {although random noise is not explicitly injected into roof masks during training, the learned model exhibits strong robustness to inherent noise in the input roof masks.}

\textbf{Down-stream validation.}
We conduct quantitative downstream evaluations on the ObliCity by comparing roof and footprint extraction across models with only image inputs. Specifically, we use an HTC ($\approx 156$M) trained only on the IRSAMap roof dataset to perform roof extraction, and then apply DragRoof to estimate RFOV. In contrast, non-decoupled baselines, \ie, LOFT [TPAMI 2023] and MLS-BRN [CVPR2024] predict roofs and RFOV, and rely on soft-NMS to retain multiple detection candidates. 
Although these methods preserve a large number of predictions (high Recall), their overall precision is relatively low.
Our decoupled formulation separates roof extraction and RFOV estimation into two independent stages, enabling more focused, task-specific training. This design likely contributes to the higher Precision and F1-score achieved by our method in the downstream roof and footprint extraction tasks (Tab.~\ref{tab:decp}).

% \subsection{Discussion}
% \subsection{SDE \vs ODE}
% \label{dis: SDE_ODE}
% % DragOSM和DragRoof都依赖于模型对图像中建筑物精准结构的感知。DragOSM的SDE训练过程会首先输入图像范围内全局的历史标签，因而模型具备建立历史标签与图上全部建筑物一一对应的能力；然而，RFOV提取则不同，模型需要对单一化输入的建筑物进行提取，然而训练于SDE假设的DragOSM在提取RFOV的过程中，周围的建筑物区域对模型带来了极大的视觉影响，将输入建筑物的屋顶与其他建筑物的footprint进行了配对，这最终导致只有第一步获取的全局RFOV是相对准确的。DragRoof建模于ODE过程，训练时模型输入的所有X_t都存在于建筑物的语义空间内部，因此本身对模型而言就已经蕴含了输入与图像建筑物之间的对应关系。
% % \textbf{Structure impact \vs training strategy. }

% \subsection{MoE \vs ODE}

% \subsection{Future}
% % off-nadir作为高分辨摄影时普遍存在的问题，随着分辨率增高越来越不容忽视，在卫星层面分辨率提高。

\section{Discussion}

\subsection{ODE Modelling for RFOV extraction}
\label{dis: SDE_ODE}

% 基于模拟人工标注过程，使用ODE建模的DragRoof在推理步数和最终RFOV精度上都处于领先地位。这种effectiveness很可能源自于图像本身所提供的几何条件路径：正如人类标注员沿着building facade拖拽roof mask的过程一样，DragRoof通过采样这条拖拽路径上的位置来模拟拖拽过程的中间状态，而给定的图像和建筑物facade则为这个拖拽提供了明确的几何路径，这很可能是DragRoof强大的方向感知的来源。相比之下，由于DragOSM的SDE训练模式假定了一个2D高斯的噪声位置空间分布，本身就少有建立输入roof到footprint一一匹配的对应关系，从而在针对RFOV的提取任务时，面对周围环绕的建筑物，无法建立对应的屋顶底座匹配关系，从而仅有第一步预测的RFOV存在有效性。
By simulating the human annotation process through ODE modeling, DragRoof achieves superior performance in both inference efficiency and RFOV accuracy. 
This effectiveness likely stems from the geometric path implicitly provided by the image itself: similar to how human annotators drag the roof mask along the building facade, DragRoof samples positions along this trajectory to mimic intermediate dragging states. 
The given image and facade structures thus provide explicit geometric guidance, which may serve as the key to its strong directional awareness.  
In contrast, DragOSM is trained under an SDE formulation that assumes a 2D Gaussian noise distribution over positional space, without establishing a one-to-one correspondence between the input roof and its footprint. 
Consequently, when surrounded by multiple adjacent buildings, the model often fails to identify the correct roof–footprint pairing, making only the first denoising step effective for valid RFOV prediction. We visualized this in Sec.~\ref{supp:ode_sde}.

% Finally, as imaging resolution continues to increase, the cost and legal complexity of producing high-precision annotations have also grown. 
% We therefore call on the research community to democratise their data resources, enabling broader access to high-quality oblique imagery for scientific exploration and methodological innovation.

\subsection{RFOV Extraction \vs Footprint Mask Metrics}
\label{supp:rfov_mask}

% 在过去，RFOV提取的相关研究常作为building footprint extraction in off-nadir remote sensing images的辅助任务，帮助获得更加准确的building footprint，因此该类型的研究主要以评估footprint mask metrics为主。而footprint mask的获得依赖于两个stage，首先通过语义分割获得roof mask，然后通过预测RFOV移动roof mask，获得建筑物的footprint mask。潜意识中，footprint mask的质量似乎与RFOV提取的质量直接相关，实则并不是这样的，Fig.~\ref{fig:insens}便列举了两种情况。
In previous studies, RFOV extraction was often treated as an auxiliary task for building footprint extraction in off-nadir remote sensing images, primarily aimed at improving footprint accuracy. 
Consequently, these works mainly evaluated performance using footprint mask metrics. 
The footprint mask itself, however, is obtained through a two-stage process: first, roof masks are produced via semantic segmentation; then, RFOVs are predicted to shift the roof masks, generating the final footprint masks. 
While it may appear that the quality of the footprint mask is directly correlated with RFOV accuracy, this is not necessarily the case. 

\begin{figure}
    \centering
    \includegraphics[width=1\linewidth]{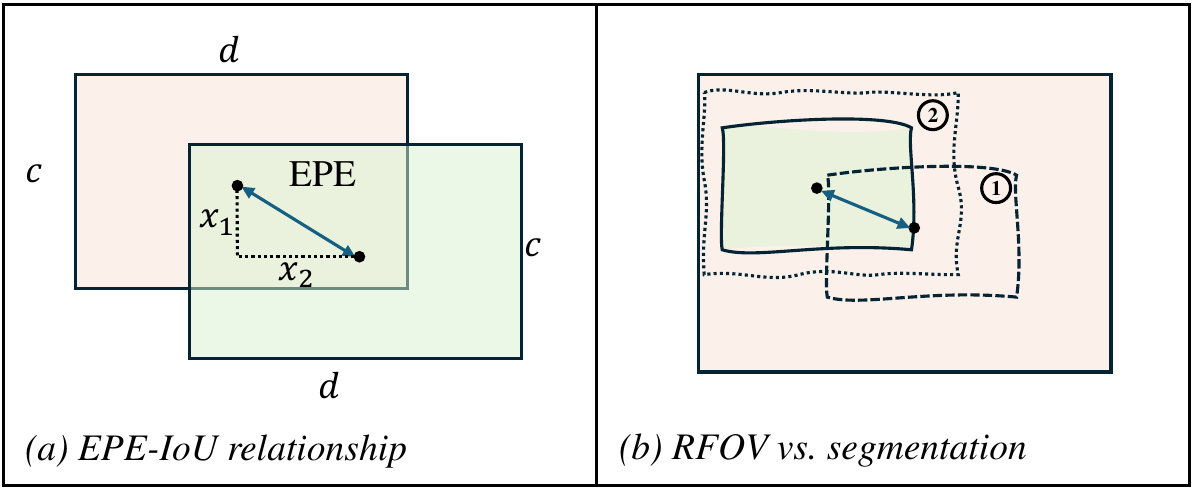}
    \caption{Two typical cases of footprint mask metrics fail to reflect RFOV improvement. }
    \label{fig:insens}
\end{figure}

As Fig.~\ref{fig:insens}, there exist cases where improvements in RFOV prediction have little to no impact on footprint mask metrics. For case Fig.~\ref{fig:insens}(a), we take footprint mask IoU as an example: assuming that the predicted and ground-truth footprint masks share identical rectangular outlines with side lengths $c$ and $d$, the current EPE can be expressed as $\text{EPE} = \sqrt{x_1^2 + x_2^2}$,
where $(x_1, x_2)$ denotes the displacement between the predicted and ground-truth footprint centers. Consequently, the footprint mask IoU is expressed as:
\begin{equation}
\label{eq:iou}
    \text{IoU} = \frac{(c-x_1)(d-x_2)}{(c+x_1)(d+x_2)-2x_1x_2}. 
\end{equation}

We fix the $c, d, x_2$ as constant in Eq.~\ref{eq:iou} to study the relationship between IoU and EPE with $x_1$, then:
\begin{align}
\label{eq:iou_2}
    \text{IoU} &= \frac{c(d-x_2)-(d-x_2)x_1}{c(d+x_2)+(d-x_2)x_1}.
    % &=\frac{c(d-x_2)}{cd+cx_2+(d-x_2)x_1}-\frac{1}{\frac{cd+cx_2}{x_1(d-x_2)}+1}. 
\end{align}
In this equation, $c > x_1$ and $d>x_2$; obviously, the EPE will drop with the decrease of $x_1$, and IoU will increase together, but the decrease of EPE and the increase of IoU were not linearly bound: when the area of the footprint mask is large enough ($c,d$ is large), the gain from the decreased EPE will be very slight. That is one of the reasons why we append ultra-long RFOV annotations with UAV images. 

Case Fig.~\ref{fig:insens}(b) is another practical problem when using footprint metrics to measure off-nadir algorithms. One of the common discoveries in predicting roof masks is that the roof prediction is always smaller than the ground-truth roof within its boundary. This made even a fully correct RFOV is provided as \textcircled{\footnotesize{1}}, the footprint mask metrics will have no change, but if we use a more advanced roof segmentation method under the same RFOV as \textcircled{\footnotesize{2}}, the footprint mask IoU will be tremendously improved. 
However, the importance of accurate RFOV extraction goes far beyond achieving precise footprint localisation, as it provides richer geometric cues for understanding building structure, height, \etc.
Moreover, in other non-ideal and unstable roof prediction cases, the impact of advances in RFOV extraction on footprint metrics becomes more complex and less predictable.

On the other hand, RFOV-bound mask metrics are highly sensitive to post-processing steps commonly used in instance segmentation, such as Non-Maximum Suppression. When a strict Non-Maximum Suppression strategy is applied to segmentation results, incorrectly removed masks will be directly reflected in the mask evaluation scores. However, this is not the case for RFOV. If predictions with longer offsets are removed, the evaluation will naturally shift toward shorter offsets, which inherently exhibit smaller errors. \eg, when the ground truth offset is 2 pixels, the error of a shorter offset prediction may fall within 1 pixel.
Such coupling fails to provide a faithful and robust assessment of RFOV extraction performance.

That is why we said that RFOV extraction has often been overshadowed by advances in roof segmentation performance, and it is time to consider it independently. 

% 在另一方面，RFOV 捆绑 mask 指标很容易受到实例分割一些后处理的算法的影响，例如，NMS。一个严格的NMS算法被apply到分割结果时，被错误删除的mask将会直接反应在mask的指标上；然而，RFOV则不然，如果删除了一些longer offset，那么评价指标自然会向shorter offset靠拢，而shorter offset天然就有较小的error，\ie, ground truth 为 2 pixels 长，error可能在 1 以内。这种捆绑无法真实而鲁棒地评估RFOV的提取效果。

\subsection{ODE \vs SDE}
\label{supp:ode_sde}
% DragOSM的提出是为了解决矢量地图与更新影像之间的不比配问题。由于历史地图中的建筑物标签之间的相对位置不明确，因此DragOSM在训练的过程中使用了2D的高斯假设去建模影像中的建筑物与历史标签之间的位置关系。而DragRoof的提出就是专门为了解决倾斜视角下的投影差问题，它依据于模拟学习人工标注员沿着建筑物facade拖拽roof的标注过程，获得校正projection difference的能力。依据于数学的抽象，DragOSM和DragRoof恰好可以被分类为基于SDE和ODE的方法。
\begin{figure}[h]
    \centering
    \includegraphics[width=1\linewidth]{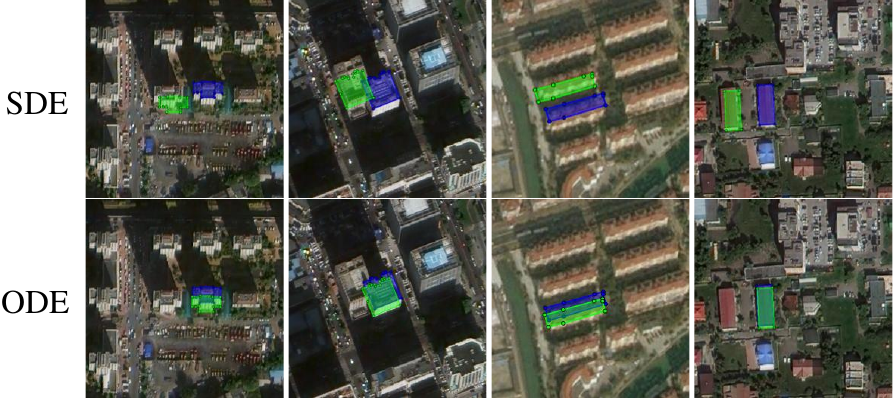}
    \caption{Comparison between DragOSM and DragRoof. The blue polygon is the location of input polygons, and the green polygons are the position of the final outputs under the predicted offset vectors. }
    \label{fig:ode_sde}
\end{figure}
DragOSM was originally proposed to address the misalignment between vector maps and newly captured imagery. 
Since the relative positions of building labels in historical maps are often uncertain, DragOSM models the spatial relationship between image buildings and historical annotations using a 2D Gaussian assumption during training. 
In contrast, DragRoof is specifically designed to handle projection differences in off-nadir imagery. 
It learns to correct such displacements by imitating how human annotators drag roofs along building facades toward their footprints. 
From a mathematical perspective, DragOSM and DragRoof can be respectively categorized as SDE-based and ODE-based formulations.

% 然而，SDE的学习在数学层面是一个维纳过程，而ODE的学习则有明确的路径。如图Fig.~\ref{fig:ode_sed},我们可视化了这两种方法的拖拽起点和终点，以辅助理解这种差异。在SDE的假设下，由于没有明确的拖拽路径，因此单一输入的roof受周围建筑物的视觉干扰，模型可能错误地估计了建筑物和标签之间的对应关系，从而被拖拽到了周围的其他建筑物，而ODE的建模利用建筑物facade本身就具备的明确轮廓关系，为inference提供了几何的路径条件，从而没有出现SDE模型的错误匹配问题。
However, learning under the SDE formulation corresponds to a Wiener process, whereas the ODE formulation follows a deterministic trajectory. 
As illustrated in Fig.~\ref{fig:ode_sde}, we visualize the starting and ending points of the dragging process to highlight this distinction. 
Under the SDE assumption, the absence of an explicit dragging path causes the input roof to be influenced by surrounding buildings, which may lead the model to misassociate the roof with incorrect footprint labels and drift toward neighboring structures. That is the reason why only the first step inference of DragOSM is reliable for extracting the RFOV. 

In contrast, the ODE formulation leverages the well-defined geometric contours along building facades to provide a clear path condition during inference, effectively eliminating such mismatched correspondences.

% \subsection{Visualization \& Generalization}
% \label{supp:visual}
% \begin{figure}
%     \centering
%     \includegraphics[width=1\linewidth]{Imgs/else_results.pdf}
%     \caption{Other results of DragRoof from the test set of ObliCity. }
%     \label{fig:othertest}
% \end{figure}

% \begin{figure}
%     \centering
%     \includegraphics[width=1\linewidth]{Imgs/zero_shot.pdf}
%     \caption{Generalizations of DragRoof on unseen data. }
%     \label{fig:zeroshot}
% \end{figure}

% % 我们在Fig.~\ref{fig:othertest}上更多地可视化了DragRoof在ObliCity数据集上的测试集的结果。DragRoof展示出良好的预测能力。同时，我们在一些未见数据集的测试集上，使用在ObliCity数据集上训练的DragRoof，进行了直接推理（Fig.~\ref{fig:zeroshot}），结果显示，我们的模型具备一定的泛化能力。

% We further visualize the qualitative results of DragRoof on the ObliCity test set in Fig.~\ref{fig:othertest}, where the model demonstrates strong predictive capability across diverse scenes. 
% In addition, we conduct zero-shot inference on several unseen datasets using the DragRoof model, which was trained solely on ObliCity (Fig.~\ref{fig:zeroshot}). 
% The results indicate that our model exhibits promising generalization ability beyond the training domain.

\subsection{Value \& Future}

% % off-nadir下的倾斜建筑物projection displacement作为高分辨摄影时普遍存在的情况，随着分辨率增高越来越不容忽视。多变的相机视角使得过去的一些建筑物方法变得逐渐不再适用，但忽视这种displacement的建筑物方法研究依然是研究的主流,而我们提出将RFOV的提取分离解耦出来，将在一定程度上更加合理化如上类型建筑物方法研究的实践可行性. 摆脱了语义任务的束缚，RFOV将可作为一个独立任务而被研究，这将极大地促进该领域的发展。最后，由于RFOV提取的数据，随着实际的摄影测量方法发展变得越来越高分辨率，同时标注的成本也越来越高，再涉及到高精度影像的开源在各个国家都存在相关法律审核，学界获得这些数据变得越来越困难，因此我们也呼吁全球的研究学者democratize自己手中的数据，供于研究。
Projection displacement in off-nadir building imagery is a common phenomenon in high-resolution remote sensing, and it becomes increasingly significant as image resolution improves. 
The diversity of camera viewing angles has rendered many traditional building extraction approaches less effective, yet most existing studies still overlook this displacement. 

We argue that decoupling RFOV extraction as an independent task provides a more principled and practical framework for modelling such geometric effects, making off-nadir building analysis more feasible and physically consistent.
Freed from the constraints of semantic segmentation, RFOV can evolve into a standalone research direction that substantially advances urban remote sensing.  

\section{Conclusion}
In this work, we formalised RFOV extraction as an independent task for correcting projection displacements in oblique remote sensing images. 
We introduced the ObliCity, the first large-scale benchmark that combines UAV and global satellite imagery with diverse resolutions and viewing angles. 
By incorporating imitation learning to emulate human annotation behaviors, we propose DragRoof, an ODE-based reformulation of DragOSM that learns deterministic, geometry-consistent offset fields, achieving state-of-the-art performance with fewer inference steps. 
We hope ObliCity and DragRoof will establish a strong foundation for future research on projection correction in monocular remote sensing imagery.

\section*{Declaration of competing interest}
The authors declare that they have no known competing financial interests or personal relationships that could have appeared to influence the work reported in this paper.

\section*{Declaration of generative AI use}
During the preparation of this manuscript, the authors used generative AI tools (\eg, ChatGPT) solely for language polishing and grammar improvement. The authors reviewed and edited the output and take full responsibility for the content of the manuscript.

\bibliographystyle{elsarticle-harv}
\bibliography{main.bib}

\end{document}